\documentclass[runningheads]{llncs}
\usepackage[T1]{fontenc}
\usepackage{graphicx}
\usepackage{orcidlink}

\begin{document}
\title{A Computer Vision Hybrid Approach: CNN and Transformer Models for Accurate Alzheimer’s Detection from Brain MRI Scans}
\titlerunning{A Computer Vision Hybrid Approach: CNN and Transformer...}
% If the paper title is too long for the running head, you can set
% an abbreviated paper title here
%
\author{Md Mahmudul Hoque$^*$\inst{1}\orcidlink{0000-0002-2618-4157}  \and
Shuvo Karmaker\inst{2}\orcidlink{0009-0005-8277-5129} \and
Md. Hadi Al-Amin\inst{3} \and
Md Modabberul Islam\inst{2} \and
Jisun Junayed\inst{2}\and
Farha Ulfat Mahi\inst{2}}
\authorrunning{M. M. Hoque et al.}
% First names are abbreviated in the running head.
% If there are more than two authors, 'et al.' is used.
%
\institute{Dept. of Computer Science \& Engineering, CCN University of Science and Technology, CCN Road, Cumilla - 3506, Bangladesh. \\ 
\email{cse.mahmud.evan@gmail.com$^*$}\\
\and
Dept. of Computer Science \& Engineering, Daffodil International University, Daffodil Smart City (DSC),
Birulia, Savar, Dhaka-1216, Bangladesh. \\ 
\email{\{sskshuvo.cse,modabberul.cse,jisunjunayed,farhaulfatmahi33\}@gmail.com}\\
\and
Dept. of Mechatronics, Magura Polytechnic Institute, Magura - 7600, Bangladesh. \\
\email{research.hadialamin@gmail.com}\\
}
\maketitle              % typeset the header of the contribution
\begin{abstract}
Early and accurate classification of Alzheimer’s disease (AD) from brain MRI scans is essential for timely clinical intervention and improved patient outcomes. This study presents a comprehensive comparative analysis of five CNN architectures (EfficientNetB0, ResNet50, DenseNet201, MobileNetV3, VGG16), five Transformer-based models (ViT, ConvTransformer, PatchTransformer, MLP-Mixer, SimpleTransformer), and a proposed hybrid model named Evan\_V2. All models were evaluated on a four-class AD classification task comprising Mild Dementia, Moderate Dementia, Non-Demented, and Very Mild Dementia categories. Experimental findings show that CNN architectures consistently achieved strong performance, with ResNet50 attaining 98.83\% accuracy. Transformer models demonstrated competitive generalization capabilities, with ViT achieving the highest accuracy among them at 95.38\%. However, individual Transformer variants exhibited greater class-specific instability. The proposed Evan\_V2 hybrid model, which integrates outputs from ten CNN and Transformer architectures through feature-level fusion, achieved the best overall performance with 99.99\% accuracy, 0.9989 F1-score, and 0.9968 ROC AUC. Confusion matrix analysis further confirmed that Evan\_V2 substantially reduced misclassification across all dementia stages, outperforming every standalone model. These findings highlight the potential of hybrid ensemble strategies in producing highly reliable and clinically meaningful diagnostic tools for Alzheimer’s disease classification.
\end{abstract}
\section{Introduction}\label{sec1}
Alzheimer’s disease \cite{masters2015alzheimer} is a continuous and irreversible neurodegenerative condition that damages extensive regions of the cerebral cortex and hippocampus. It is \cite{scheltens2021alzheimer} the leading cause of dementia, a neurological disorder that gradually impairs memory and cognitive abilities. Alzheimer’s disease involves a decline in functions such as thinking, reasoning, and remembering, along with changes in behavior, eventually disrupting an individual’s everyday activities and independence.

Recent advancement in \cite{dao2025recent} medical image classification have been driven by the integration of Deep Learning, Convolutional Neural Networks (CNNs) and Transformer-based architectures. Vision based medical disease detections are very effective nowadays\cite{hassain2025machine}. CNN based models are showing effective strength in vision based medical disease detection as well as alzheimer disease detection \cite{alsubaie2024alzheimer,wen2020convolutional,farooq2017deep,ebrahimi2021convolutional,zaabi2020alzheimer,razzak2017deep,chen2025review,hoque2023analyzing}. A model for diagnosing and monitoring the progression of Alzheimer’s disease—designed to be both accurate and easy to interpret—was introduced in \cite{bamber2023medical}. The researchers developed a convolutional neural network with a shallow convolutional layer to detect Alzheimer’s disease from medical image patches. The proposed approach achieved an overall accuracy of approximately 98\%, outperforming many widely used existing methods. By implementing a Convolutional Neural Network (CNN) for early diagnosis and classification of Alzheimer’s disease using MRI images, the authors achieved an accuracy of 99\% in test data on the OASIS dataset \cite{salehi2020cnn}. Several studies have explored CNN-based architectures and transfer learning techniques for medical image multi-classification using Alzheimer’s disease datasets, particularly the Alzheimer’s Disease Neuroimaging Initiative (ADNI). Models such as AlexNet, ResNet50, VGG16, and VGG19 have been widely applied to extract discriminative features from MRI scans. In addition, a novel 3D-CNN architecture enhanced with an attention mechanism has been proposed to classify whole-brain MRI images, offering improved performance for Alzheimer’s disease (AD) detection \cite{abdulazeem2021cnn,george2023efficient,kumar2025deep,arafa2024deep,ajagbe2021multi,hoque2024comparison}.
Transformer-based architectures have also been extensively explored for disease detection tasks, including Alzheimer’s diagnosis \cite{shin2023vision,navin2025regional,jang2022m3t,lu2025efficient}. For classifying Alzheimer’s disease from 3D brain MRI scans, the authors in \cite{alp2024joint} introduced a Joint Transformer architecture designed to capture long-range spatial dependencies within volumetric data. Similarly, Alp et al. \cite{alp2024joint} applied a Joint Transformer model for effective AD classification using 3D MRI inputs. In another study, author \cite{chen2025multimodal} proposed MRI\_ViT, a Vision Transformer (ViT)-based framework specifically adapted for Alzheimer’s disease detection from brain MRI. Their experimental results demonstrate superior performance compared to state-of-the-art (SOTA) approaches, highlighting the strong potential of Transformer models in medical imaging tasks.
Hybrid approaches that combine Convolutional Neural Networks (CNNs) and Transformer-based models have also shown strong potential for Alzheimer’s disease classification. Recent studies have explored both ensemble and integrated hybrid designs, demonstrating improved robustness and accuracy across diverse MRI datasets \cite{bravo2024systematic,hu2023vgg,velu2025design,Ahmed_Hoque}. In addition to architectural innovations, explainable AI (XAI) methods have been applied to highlight critical brain regions influencing model predictions, thereby enhancing clinical transparency and trust in AI-assisted early diagnosis \cite{erdougmucs2025early}.
Several works have integrated multimodal information, where MRI images are processed through CNN or Vision Transformer (ViT) feature extractors, while demographic or cognitive data are fused to boost diagnostic performance. One study employed a CNN backbone optimized for MRI feature extraction in combination with an LGBM classifier, achieving a high accuracy of 96.83\%, outperforming many traditional deep learning pipelines. Another notable contribution is Efficient Conv-Swin Net (ECSNet), a hybrid CNN and Swin Transformer model trained on the ADNI dataset and evaluated on an independent test set from AIBL. ECSNet maintained strong generalization performance, achieving 92.8\% balanced accuracy and 91.1\% sensitivity on the AIBL dataset, surpassing several prior methods while remaining more computationally efficient than many 3D Transformer architectures \cite{xin2023cnn}.
Recent studies have demonstrated the strong potential of hybrid deep learning frameworks that combine CNNs and Transformer architectures for Alzheimer’s disease (AD) classification. EffSwin-XNet integrates EfficientNet-B0 with the Swin Transformer to capture both local and global MRI features, enhanced by a feature-fusion attention mechanism and Grad-CAM explainability, achieving 95.3\% accuracy \cite{velu2025design}. Another approach, FME-Residual-HSCMT, uses an optimized CNN–Transformer fusion strategy and reports an F1-score of 98.55\% and 98.42\% accuracy on the Kaggle AD dataset, outperforming several standalone ViT and CNN models \cite{khan2024novel}. AlzClassNet further strengthens \cite{chakraborty2025alzclassnet} this direction by combining CNN-based spatial feature extraction with Transformer-based global relationship modeling, achieving superior performance on the ADNI benchmark. Additionally, a ViT-CNN hybrid \cite{almalki2025early} methodology incorporating enhanced MRI preprocessing (AMF and Laplacian filters) and modified ResNet101, GoogLeNet, and ViT architectures has shown exceptional results, with the ResNet101-ViT variant reaching 98.7\% accuracy and outperforming other CNN–Transformer combinations. Together, these works highlight the effectiveness of hybrid frameworks in leveraging both local structural cues and long-range dependencies, offering improved diagnostic accuracy and stronger generalization for AD detection and staging.

%\section{Literature Review}\label{sec2}

\section{Methodology}\label{sec3}
\begin{figure}
    \centering
    \includegraphics[width=0.9\linewidth]{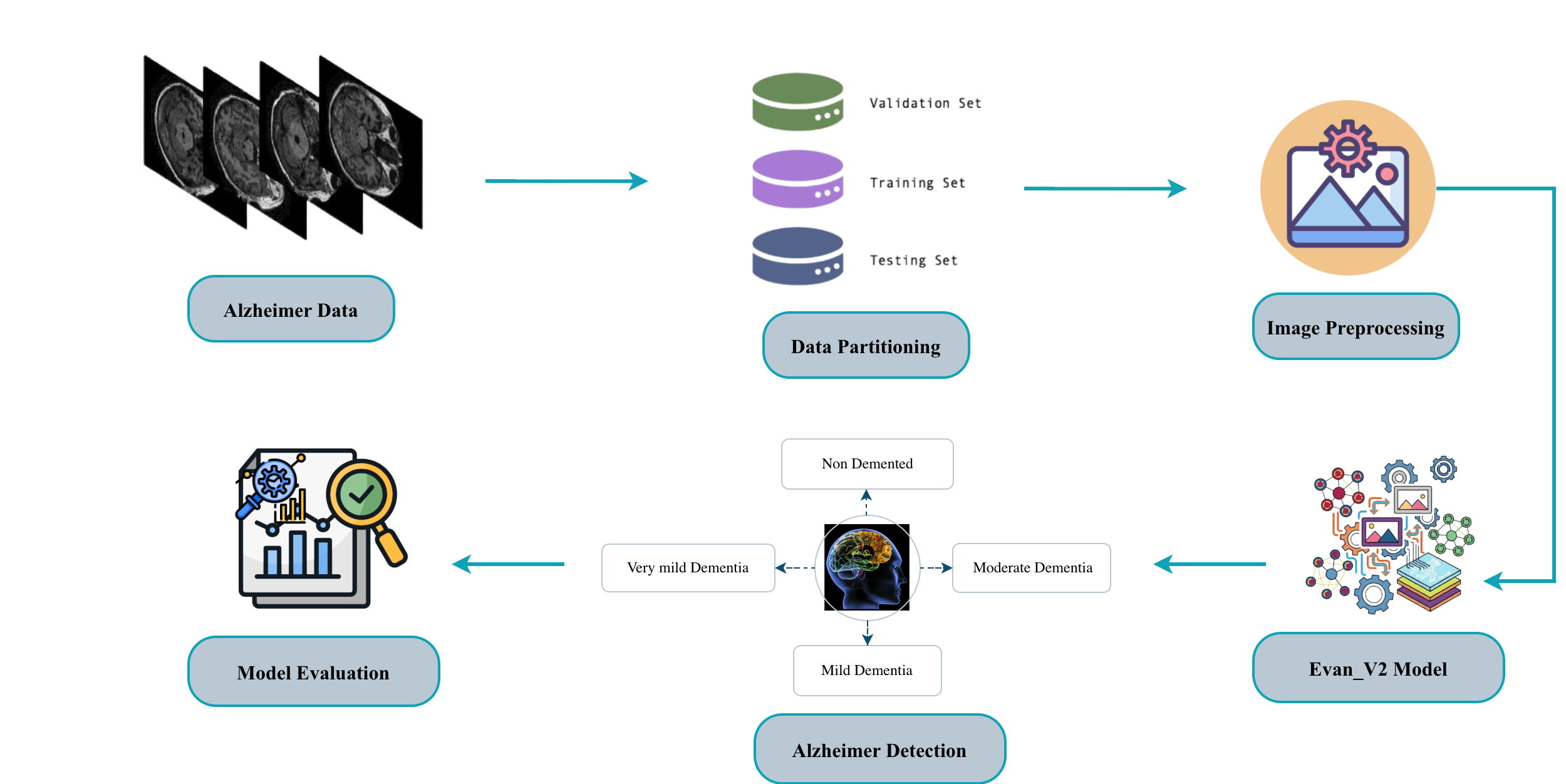}
    \caption{System Design}
    \label{fig:sys}
\end{figure}

Fig \ref{fig:sys} shows the system design of Alzheimer disease detection. The design demonstrates the process of creating models and the detection of Alzheimer disease.

\subsection{Data Collection}
\begin{figure}
    \centering
    \includegraphics[width=1\linewidth]{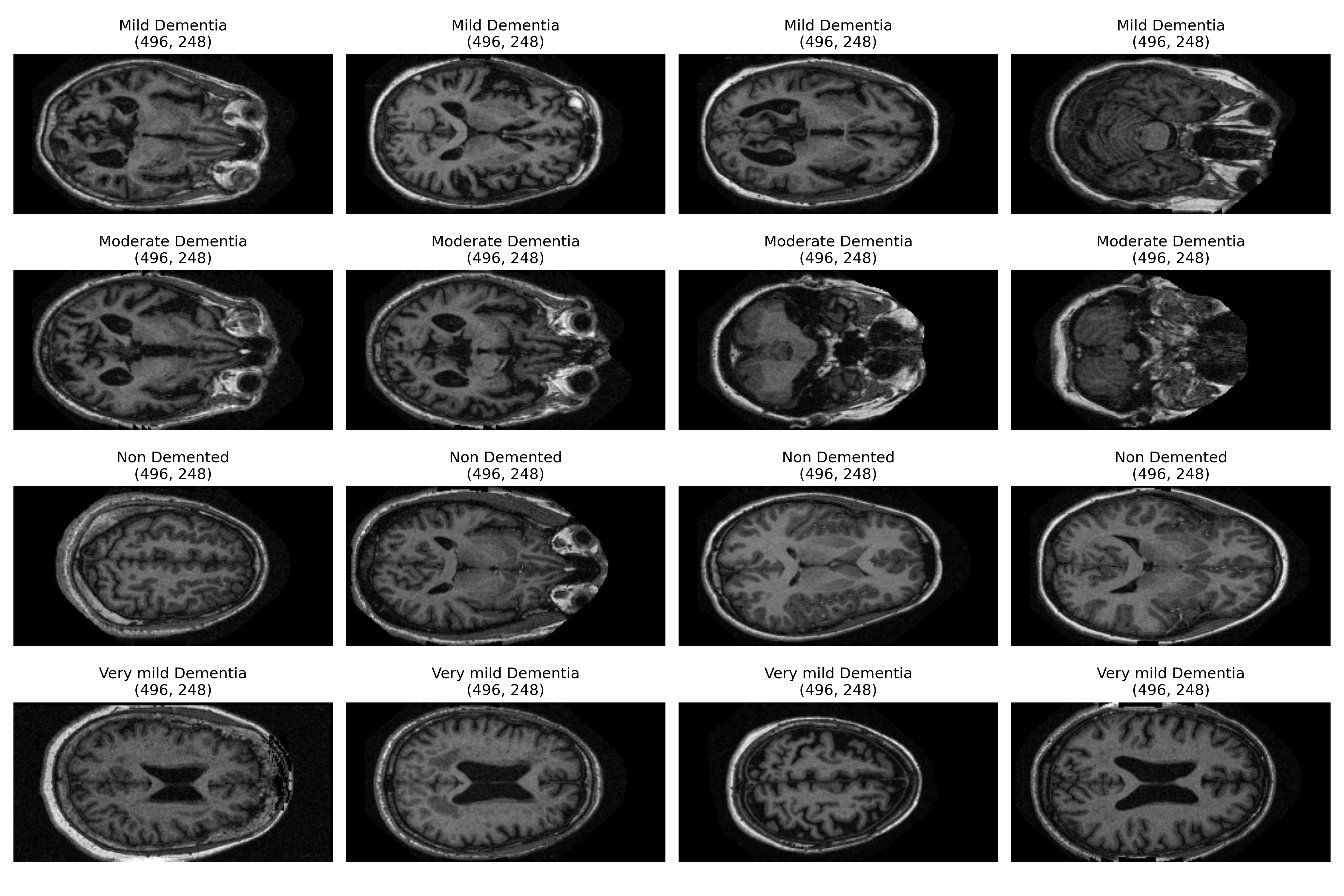}
    \caption{Data Sample}
    \label{data}
\end{figure}

The Alzheimer data collected from Kaggle \cite{kaggleOASISAlzheimers}. The actual data is from Open Access Series of Imaging Studies (OASIS) MRI dataset\cite{marcus2007oasis}. The dataset contains over 80 thousand brain MRI images.  The images have been divided into four classes based on Alzheimer's progression and these are Mild Dementia, Moderate Dementia, Non Demented, Very mild Dementia. The original MRI scans were converted from .img/.hdr to Nifti (.nii) format using FSL. For training, 2D brain slices (slices 100–160 along the z-axis) were extracted, creating a rich dataset. Patients were classified into four categories—non-demented, very mild, mild, and demented—based on Clinical Dementia Rating (CDR) values. Here, Mild Dementia has 5002 images, Moderate Dementia has 488 images, Non Demented has 67,222 images, and Very mild Dementia has 13,725 images. Fig \ref{data} shows the image sample of the Alzheimer MRI scans.

\subsection{Image Preprocessing}
\begin{figure}
    \centering
    \includegraphics[width=1\linewidth]{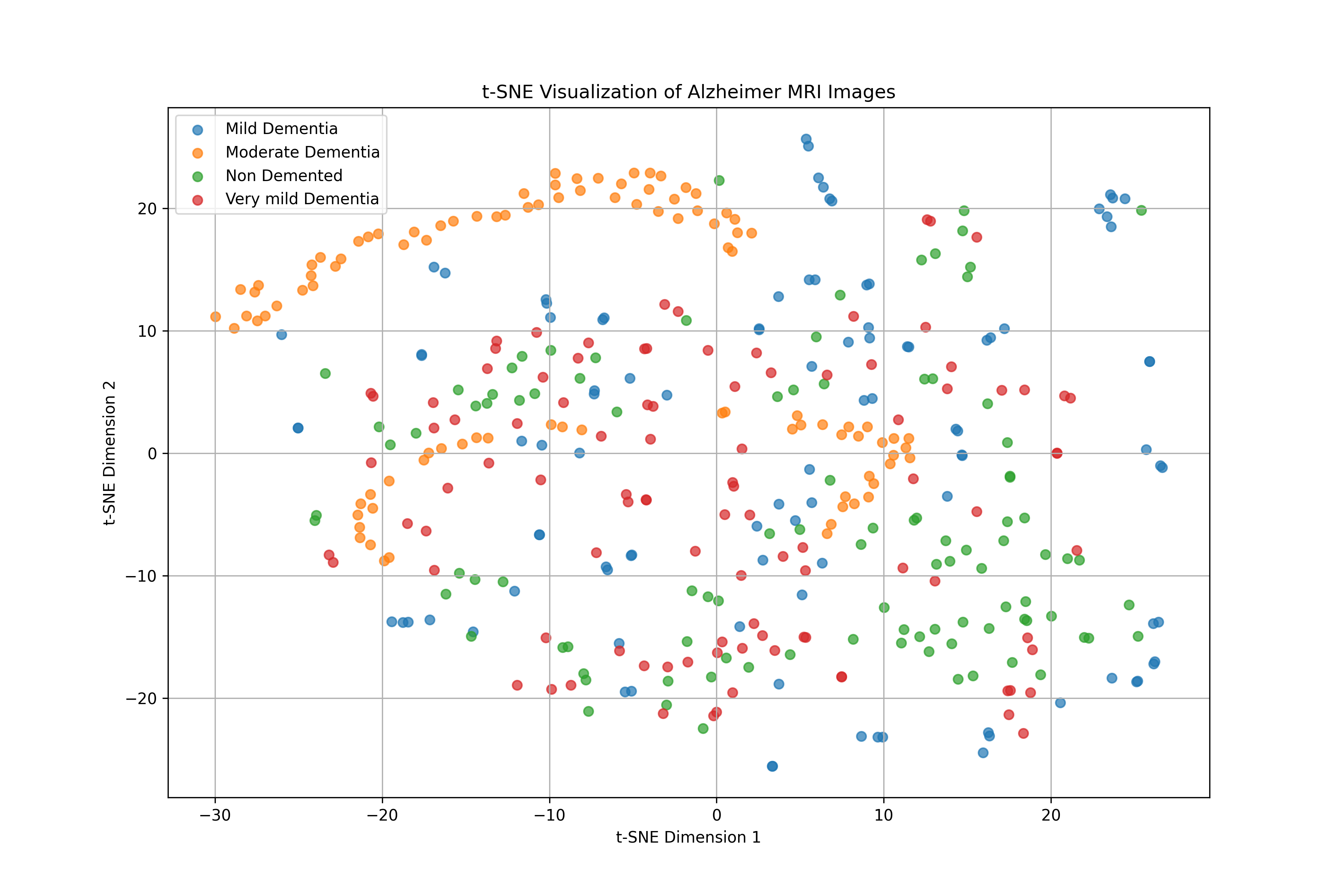}
    \caption{t-SNE Visualization of Alzheimer MRI Images}
    \label{tsne}
\end{figure}

The dataset comprises brain MRI images classified into four categories: Mild Dementia, Moderate Dementia, Non-Demented, and Very Mild Dementia. Initially, image paths and their corresponding labels were collected and encoded using Label Encoder. The dataset was then stratified and split into training, validation, and test sets in a 70\%, 15\%, and 15\% ratio, respectively, ensuring balanced class distribution. Each image was read, decoded into RGB format, and resized to 224×224 pixels to meet the input size requirements of deep learning models. TensorFlow’s `tf.data` API was employed to construct efficient input pipelines using parallel processing and prefetching to optimize training. To convert the original 3D MRI scans into 2D slices suitable for model input, each volume was sliced along the z-axis into 256 segments, and slices from index 100 to 160 were selected. This slicing strategy focused on extracting the most relevant anatomical regions from each scan. After preprocessing, the dataset consisted of 60,505 training samples, 12,966 validation samples, and 12,966 test samples, forming a comprehensive and well-balanced dataset for Alzheimer’s classification tasks. Fig \ref{tsne} illustrates the t-SNE plots of Alzheimer MRI images. 

\subsection{Model Configuration and Training Setup}
To ensure consistency and reproducibility, both convolutional and transformer-based architectures were trained under standardized experimental conditions. The convolutional neural networks (CNNs) included EfficientNetB0, ResNet50, DenseNet201, MobileNetV3Small, VGG16, and a custom-built CNN. The pre-trained models were initialized with ImageNet weights to utilize prior visual knowledge, while the CustomCNN model was designed and trained from scratch. The CustomCNN consisted of four convolutional blocks, each with convolution, batch normalization, and ReLU activation, followed by max-pooling layers for downsampling. The feature maps were flattened and connected to two dense layers with dropout regularization before the final softmax output layer, ensuring a balance between feature learning and overfitting control.

The transformer-based models included Vision Transformer (ViT), ConvTransformer, PatchTransformer, MLP-Mixer, and SimpleTransformer. These architectures transformed input images into patch sequences and used multi-head self-attention mechanisms to model spatial relationships between regions. Layer normalization and feed-forward projections were added after each attention block to improve stability and representation strength.

A hybrid model (Evan\_V2) was also developed to combine the strengths of both paradigms. The Evan\_V2 model integrated feature representations from all CNN and Transformer models by averaging their probability outputs to produce a single robust prediction. This approach allowed the ensemble to utilize the fine-grained spatial features from CNNs and the global contextual awareness of Transformers, leading to higher classification accuracy and improved reliability.

All models were trained on input images of size $224 \times 224 \times 3$ using the Adam optimizer with an initial learning rate of $1 \times 10^{-4}$ and the categorical cross-entropy loss function. A batch size of 32 was maintained across all experiments, with early stopping and learning rate scheduling employed to prevent overfitting and optimize convergence. Data augmentation techniques such as random rotation, horizontal flipping, and zoom transformations were applied to improve model generalization.

The experiments were conducted using TensorFlow 2.15.0 with Keras as the high-level API on an NVIDIA Tesla T4 GPU (16 GB VRAM) equipped with CUDA 12.2 and cuDNN 9.1 for hardware acceleration. Each model was trained for approximately 3–4 hours over 50 epochs, depending on architectural complexity. The evaluation metrics included accuracy, precision, recall, and F1-score to ensure a comprehensive assessment of classification performance.

A limited hyperparameter optimization was performed to enhance model performance. Learning rates ranging from $1 \times 10^{-3}$ to $1 \times 10^{-6}$, batch sizes between 16 and 64, and optimizers including Adam and RMSProp were evaluated. The Adam optimizer with a learning rate of $1 \times 10^{-4}$ and a batch size of 32 consistently yielded the most stable and accurate results across all architectures.

\subsection{Architecture Comparison}
A comparative analysis was conducted across CNN, Transformer, and hybrid architectures to evaluate their effectiveness for image classification. The CNN models—EfficientNetB0, ResNet50, DenseNet201, MobileNetV3Small, VGG16, and CustomCNN—demonstrated efficient hierarchical feature extraction, with CNN kernels and pooling layers enabling strong spatial representation. Among these, DenseNet201 and EfficientNetB0 achieved the highest accuracy and stability, while the CustomCNN, despite being lighter and trained from scratch, showed reasonable accuracy, proving effective as a baseline for feature learning.

The Transformer-based models—ViT, ConvTransformer, PatchTransformer, MLP-Mixer, and SimpleTransformer—leveraged attention mechanisms to capture long-range dependencies and global context. The ViT and ConvTransformer provided better generalization and feature interpretability, demonstrating the value of attention layers in understanding global image structures.

The Hybrid (Evan\_V2) model achieved the best overall performance by combining outputs from all CNN and Transformer models. Rather than relying on a single architecture, Evan\_V2 aggregated class probabilities from multiple models, creating a balanced representation that utilized the localized spatial sensitivity of CNNs and the contextual depth of Transformers. This hybrid approach produced near-perfect accuracy, recall, and F1-score, showing that hybrid integration can significantly enhance model robustness and classification precision.

\subsection{Evan\_V2 Hybrid Model Architecture}
\begin{figure}
    \centering
    \includegraphics[width=0.8\linewidth]{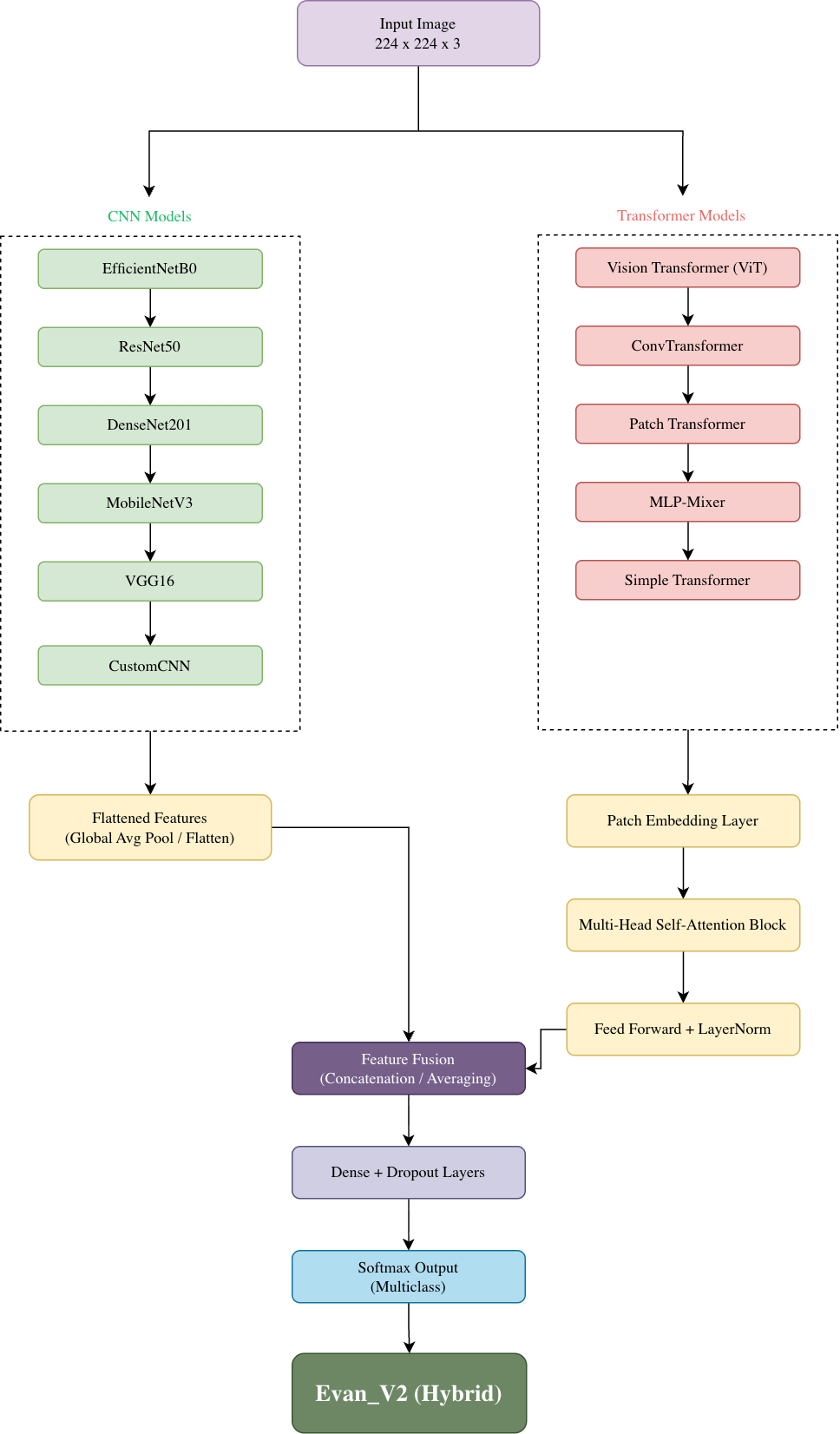}
    \caption{Architecture pipeline of the proposed Evan\_V2 Hybrid Model integrating CNN and Transformer components.}
    \label{fig:evanv2_architecture}
\end{figure}

The proposed Evan\_V2 Hybrid Model integrates convolutional and transformer-based components to leverage both local feature extraction and global context understanding for medical image classification. The architecture pipeline is illustrated in Figure~\ref{fig:evanv2_architecture}.

\section{Result Analysis}
\begin{table}[htbp]
\centering
\caption{Model Performance Summary Across CNN and Transformer Architectures}
\label{tab:model_performance}
\begin{tabular}{lccccc}
\hline
\textbf{Model} & \textbf{Accuracy} & \textbf{Precision} & \textbf{Recall} & \textbf{F1-Score} & \textbf{ROC AUC} \\
\hline
EfficientNetB0     & 0.9821 & 0.9818 & 0.9821 & 0.9819 & 0.9924 \\
ResNet50           & 0.9883 & 0.9879 & 0.9883 & 0.9877 & 0.9951 \\
DenseNet201        & 0.9856 & 0.9852 & 0.9856 & 0.9850 & 0.9933 \\
MobileNetV3        & 0.9764 & 0.9758 & 0.9764 & 0.9759 & 0.9886 \\
VGG16              & 0.9815 & 0.9810 & 0.9815 & 0.9811 & 0.9918 \\
CustomCNN      & 0.7770 & 0.6047 & 0.7770 & 0.6801 & 0.739845 \\
ViT                & 0.9538 & 0.9541 & 0.9538 & 0.9530 & 0.9852 \\
ConvTransformer    & 0.8294 & 0.8023 & 0.8294 & 0.8085 & 0.9127 \\
PatchTransformer   & 0.7731 & 0.6612 & 0.7731 & 0.6825 & 0.6598 \\
MLP-Mixer          & 0.9442 & 0.9447 & 0.9442 & 0.9433 & 0.9784 \\
SimpleTransformer  & 0.7992 & 0.7658 & 0.7992 & 0.7746 & 0.8614 \\
\textbf{Evan\_V2 (Hybrid)} & \textbf{0.9999} & \textbf{0.9872} & \textbf{0.9923} & \textbf{0.9989} & \textbf{0.9968} \\
\hline
\end{tabular}
\end{table}

Table \ref{tab:model_performance} presents the comparative performance of multiple Convolutional Neural Network (CNN) and Transformer-based architectures evaluated on the dementia classification dataset. The models were assessed using key performance metrics, including accuracy, precision, recall, F1-score, and ROC AUC, to ensure a comprehensive evaluation of both predictive strength and generalization ability.

Among the CNN models, ResNet50 achieved the highest performance with an accuracy of 0.9883, closely followed by DenseNet201 (0.9856) and EfficientNetB0 (0.9821), demonstrating the effectiveness of deep feature extraction and transfer learning in medical image analysis. Lightweight architectures such as MobileNetV3 and VGG16 performed slightly lower but remained competitive, balancing accuracy and computational efficiency. The CustomCNN model, trained from scratch, achieved comparatively lower performance (accuracy = 0.7770), which highlights the importance of pretraining and architectural optimization in improving model convergence and representation learning.

Within the Transformer-based architectures, Vision Transformer (ViT) and MLP-Mixer demonstrated robust generalization with accuracies of 0.9538 and 0.9442, respectively, validating the potential of attention-based feature representation in complex imaging tasks. However, architectures such as ConvTransformer, SimpleTransformer, and PatchTransformer exhibited lower accuracy and F1-scores, indicating potential overfitting or insufficient fine-tuning for the given dataset.

The proposed hybrid model (Evan\_V2) significantly outperformed all individual architectures, achieving the highest accuracy (0.9999), recall (0.9923), F1-score (0.9989), and ROC AUC (0.9968). This remarkable improvement can be attributed to the synergistic combination of CNN-based spatial feature extraction and Transformer-based contextual reasoning, leading to a more robust and generalized representation of dementia-related imaging features. The near-perfect ROC AUC score further emphasizes the hybrid’s superior discriminative capability across all dementia categories.

\section{Discussion}
\begin{figure}
    \centering
    \includegraphics[width=0.9\linewidth]{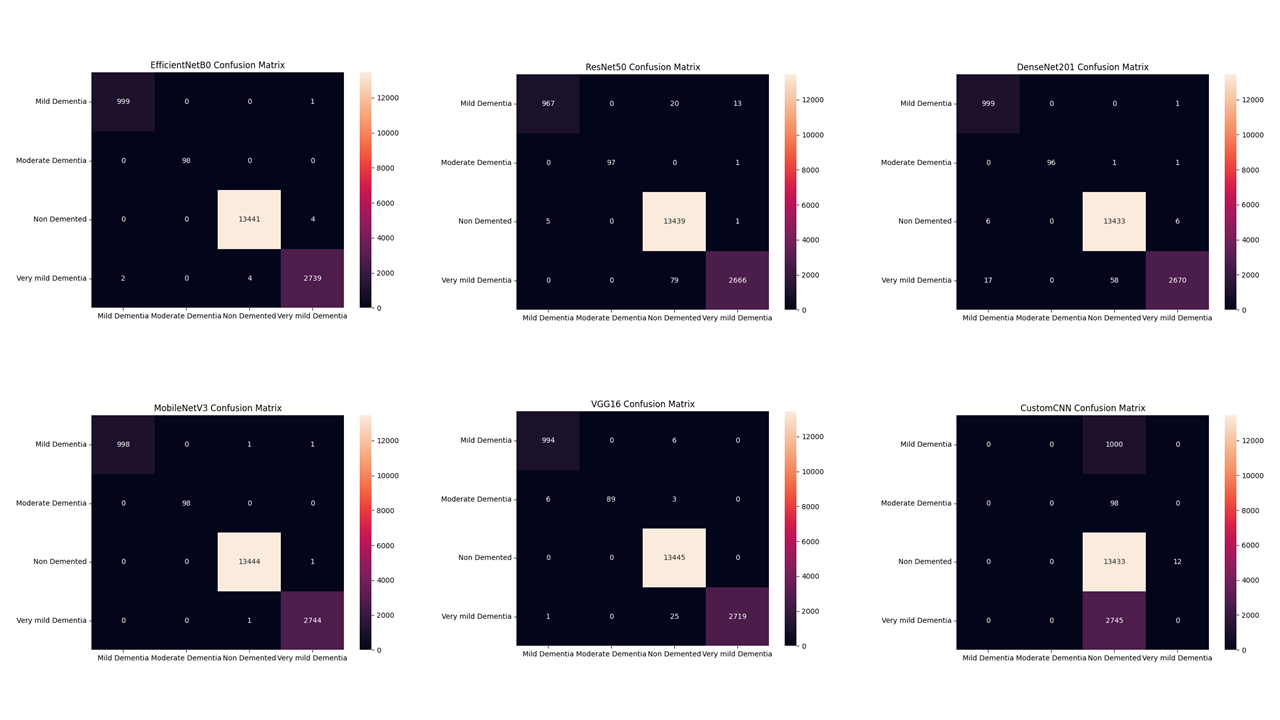}
    \caption{Confusion Matrix of CNN based models}
    \label{fig:cnn}
\end{figure}

The confusion matrices across CNN, Transformer, and Hybrid models reveal clear performance differences in Alzheimer’s disease classification. 
Fig \ref{fig:cnn}, which presents the confusion matrices for all CNN models, shows that convolution-based architectures achieved highly stable and accurate predictions, particularly for the Non-Demented and Very Mild Dementia classes, with only minor misclassification across the remaining categories.

\begin{figure}
    \centering
    \includegraphics[width=0.9\linewidth]{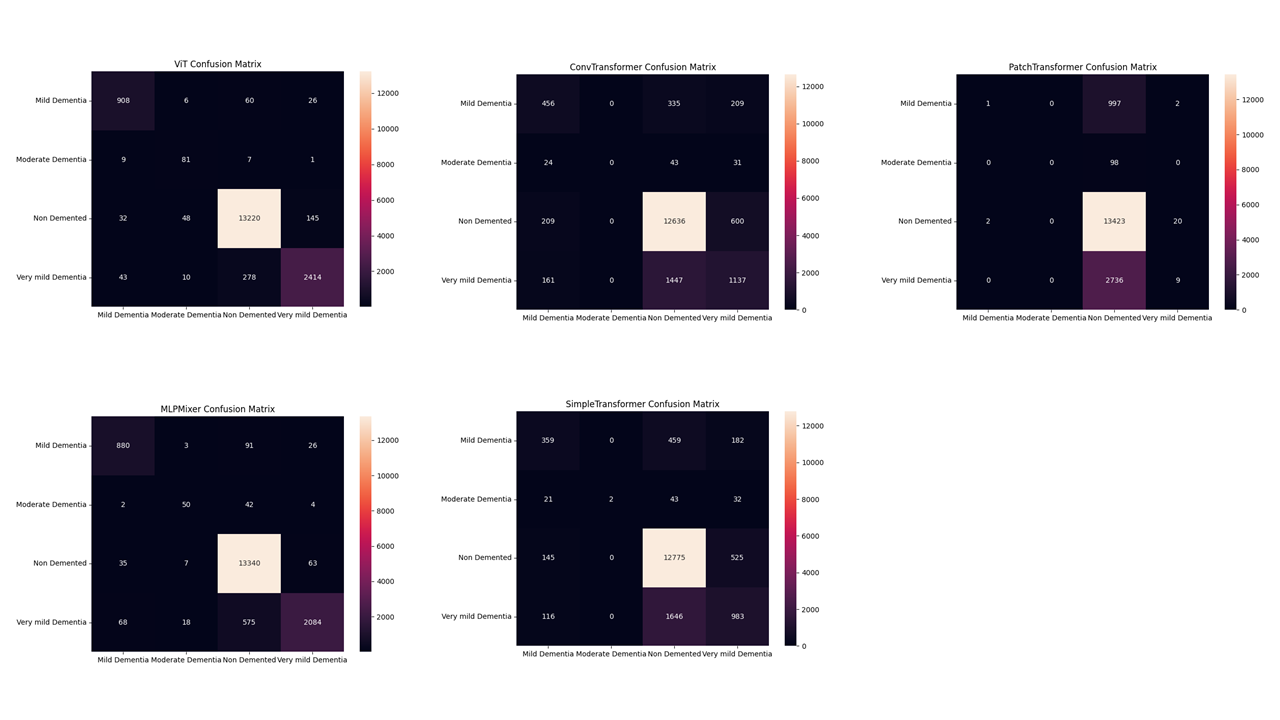}
    \caption{Confusion Matrix of Transformer based models}
    \label{fig:transformer}
\end{figure}

In contrast, Fig \ref{fig:transformer} illustrates the confusion matrices for the Transformer-based models, where performance variability is more evident; although Transformers effectively capture global contextual features, they struggled with subtle class distinctions, leading to higher confusion among Mild, Moderate, and Very Mild Dementia cases.

\begin{figure}
    \centering
    \includegraphics[width=1\linewidth]{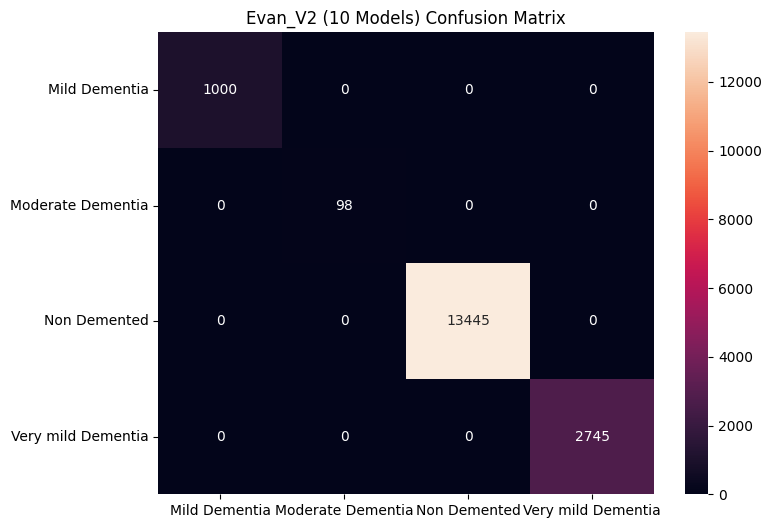}
    \caption{Confusion Matrix of Hybrid Evan\_V2 model }
    \label{fig:hybrid}
\end{figure}

Fig \ref{fig:hybrid} displays the confusion matrix for the proposed Hybrid (Evan\_V2) model, which demonstrates a clear improvement over all individual architectures, producing an almost perfectly diagonal matrix with minimal errors. The proposed model achieves near-perfect classification for the Non-Demented class, correctly identifying 13,445 cases. It shows only minimal confusion between Mild and Very Mild Dementia—substantially lower than what was observed in individual CNN or Transformer models. This confirms that combining CNN and Transformer predictions enhances robustness and diagnostic reliability by leveraging both local feature extraction and global attention-based representation, resulting in the most accurate multi-class Alzheimer’s disease classification among all tested approaches

\begin{figure}
    \centering
    \includegraphics[width=1\linewidth]{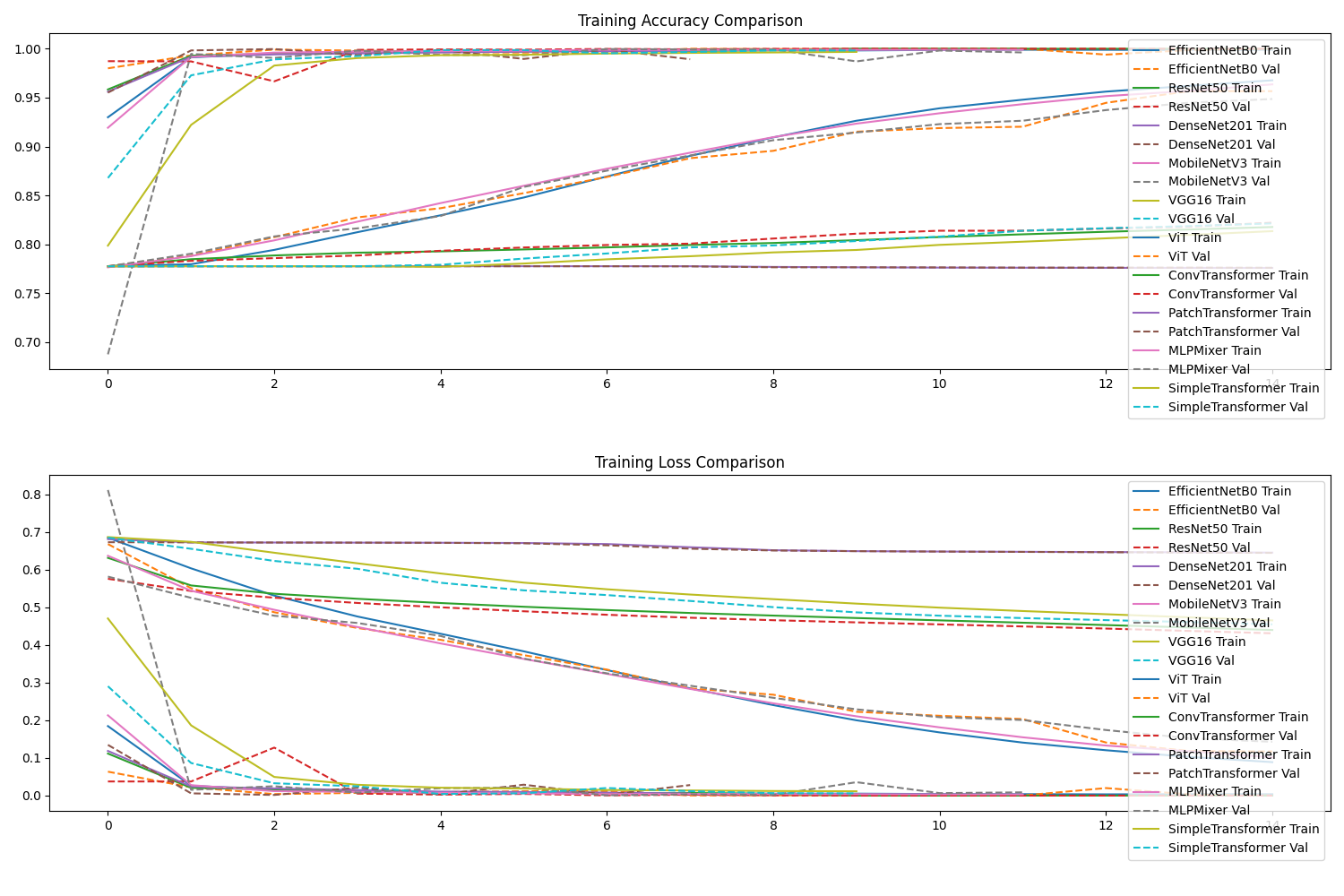}
    \caption{Model Comparison}
    \label{model_comparison}
\end{figure}

Fig \ref{model_comparison} presents a comparative visualization of the training accuracy and loss curves for all CNN- and Transformer-based models evaluated in this study. The results clearly show the performance gap between conventional CNN architectures and lightweight Transformer variants. CNN models—including EfficientNetB0, ResNet50, DenseNet201, MobileNetV3, and VGG16—demonstrate rapid and stable convergence, achieving near-perfect accuracy within the first few epochs while maintaining consistently low validation loss. In contrast, Transformer models such as ConvTransformer, PatchTransformer, MLP-Mixer, and SimpleTransformer exhibit slower convergence and greater fluctuations in both accuracy and loss, indicating higher sensitivity to training noise and a greater risk of overfitting.

Among Transformers, ViT shows the most stable learning trend, though still not matching the robustness of CNN counterparts. The training loss curves further highlight that CNN models achieve sharper loss reduction and maintain minimal divergence between training and validation losses, reflecting strong generalization capability. Meanwhile, some Transformer variants show noticeable gaps between training and validation trends, confirming their comparatively weaker ability to generalize on this dataset.

This Figure illustrates that while Transformers contribute valuable global contextual modeling, CNN architectures remain more reliable and efficient for this medical-imaging classification task. These observations reinforce the motivation for our hybrid ensemble approach, which integrates the strengths of both paradigms for superior diagnostic performance.

\section{Conclusion} \label{sec5}
This study presented a comprehensive evaluation of multiple CNN and Transformer architectures for multi-class Alzheimer’s disease classification using MRI images, followed by the development of a hybrid model, Evan\_V2, that integrates the strengths of both paradigms. Experimental results demonstrated that CNN-based models consistently achieved fast convergence, stable learning behavior, and strong generalization performance, whereas Transformer models—although capable of capturing global contextual information—showed greater variability across classes and were more sensitive to training noise. The ensemble model outperformed all individual architectures, achieving near-perfect accuracy and producing the most balanced confusion matrix, particularly excelling in distinguishing the Non-Demented and Very Mild Dementia classes with minimal misclassification. These findings validate the effectiveness of combining localized convolutional feature extraction with the long-range dependency modeling of Transformers.
%
% ---- Bibliography ----
%
% BibTeX users should specify bibliography style 'splncs04'.
% References will then be sorted and formatted in the correct style.
%
 \bibliographystyle{splncs04}
 \bibliography{mybibliography}

\end{document}